\documentclass[journal]{IEEEtran}

\usepackage[english]{babel}

\usepackage{amsmath}
\usepackage{lipsum} 
\usepackage[T1]{fontenc}
\usepackage{newtxtext}
\usepackage[cmintegrals]{newtxmath}
\usepackage{bm} 
\usepackage[utf8]{inputenc}
\usepackage{graphicx}
\usepackage{graphics}

\usepackage{siunitx}
\usepackage{multirow}
\usepackage{float}

\usepackage{gensymb}
\usepackage{hyperref}
\usepackage{epstopdf}
\usepackage{enumerate}
\usepackage{multicol,multirow}
\usepackage{booktabs}
\usepackage{cite}
\usepackage{xcolor}

\usepackage[ruled,lined,linesnumbered]{algorithm2e}
\let\oldnl\nl
\newcommand{\nonl}{\renewcommand{\nl}{\let\nl\oldnl}}

\graphicspath{{figures/}}
\usepackage{balance}
\usepackage{xcolor}
\newcommand\SelCol{black}
\begin{document}

	\title{\textsc{HapFIC}: An Adaptive Force/Position Controller for Safe Environment Interaction in Articulated Systems}
	\author{Carlo~Tiseo, Wolfgang~Merkt, Keyhan~Kouhkiloui~Babarahmati, Wouter~Wolfslag, Ioannis~Havoutis, Sethu~Vijayakumar and Michael~Mistry
		\thanks{Carlo Tiseo, Keyhan Kouhkiloui Babarahmati,Wouter Wolfslag,  Sethu~Vijayakumar and Michael Mistry are with the School of Informatics, University of Edinburgh.
		Wolfgang~Merkt and Ioannis~Havoutis are with the Oxford Robotics Institute, University of Oxford, Oxford, England, UK. Email: \texttt{carlo.tiseo@ed.ac.uk}.}
	}
	\maketitle
	
	\begin{abstract}
        Haptic interaction is essential for the dynamic dexterity of animals, which seamlessly switch from an impedance to an admittance behaviour using the force feedback from their proprioception. However, this ability is extremely challenging to reproduce in robots, especially when dealing with complex interaction dynamics, distributed contacts, and contact switching. Current model-based controllers require accurate interaction modelling to account for contacts and stabilise the interaction. In this manuscript, we propose an adaptive force/position controller that exploits the fractal impedance controller's passivity and non-linearity to execute a finite search algorithm using the force feedback signal from the sensor at the end-effector. The method is computationally inexpensive, opening the possibility to deal with distributed contacts in the future. We evaluated the architecture in physics simulation and showed that the controller can robustly control the interaction with objects of different dynamics without violating the maximum allowable target forces or causing numerical instability even for very rigid objects. The proposed controller can also autonomously deal with contact switching and may find application in multiple fields such as legged locomotion, rehabilitation and assistive robotics.
	\end{abstract}
 	
	
	\IEEEpeerreviewmaketitle

\section{Introduction} \label{sec:Intro}
The interaction skills of animals in unstructured environments are possible thanks to their ability to safely interact with unknown and complex dynamics in their daily activities. Examples include activities that involve interacting with soft objects, handling fluids, or walking in a crowded room. If we look at these tasks in the context of robotics, they all continue to be open research questions \cite{Chatzinikolaidis2020,xin2020,manchester2020,dong2019adaptive,Nah2020,Averta2020,tiseo2020,ferrolho2020}. The methods currently deployed rely on accurate environmental interaction models that might require tracking non-accessible environmental states. Notwithstanding the modelling challenge, the curse-of-dimensionality makes them computationally intensive for higher-dimensional systems \cite{manchester2020,Chatzinikolaidis2020,Stouraitis2020,HumanLikeAdaptOfForceAndImp}. The feasibility of available architectures so far has focused on small scale scenarios with controlled interaction conditions (e.g. known contact dynamics). These optimisation algorithms and controllers also exhibited a lack of robustness, which is connected to the need for accurate task models to guarantee interaction stability \cite{xin2020,Averta2020,manchester2020,Chatzinikolaidis2020}. Improving both interaction robustness and haptics is of interest for robotics at large. Furthermore, it is essential in rehabilitation and assistive robotics where our technologies are closely interacting with frail subjects. Therefore, the generation of realistic virtual environments is critical for delivering effective therapy using robots exploiting both virtual and enhanced reality. Better haptic controllers can help provide a more natural interaction to the users.

\begin{figure}[t]
    \centering
    \includegraphics[width=0.49\linewidth,trim={9cm 3cm 4cm 5cm},clip]{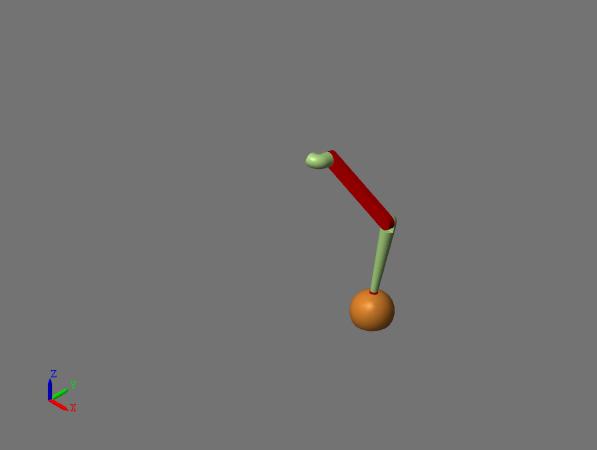}
    \hfill
    \includegraphics[width=0.49\linewidth,trim={5.5cm 3cm 6.5cm 4cm},clip]{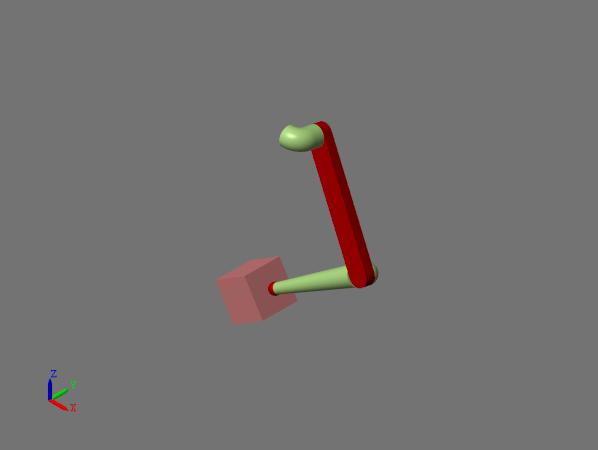}
    \caption{
    We evaluate our method, \textsc{HapFIC}, in scenarios interacting with a squishy ball (left) and rigid box (right) in a full dynamics simulation on a 3-DoF anthropomorphic manipulator/quadruped leg.
    Our experiments demonstrate that it is able to stabilise contacts in absence of friction and exhibit favourable performance in interactions with squishy and rigid objects.
    }
    \label{fig:intro}
\end{figure}

\begin{figure*}[t]
    \centering
    \includegraphics[width=0.8\linewidth,trim={0.8cm 6.2cm 0 0},clip]{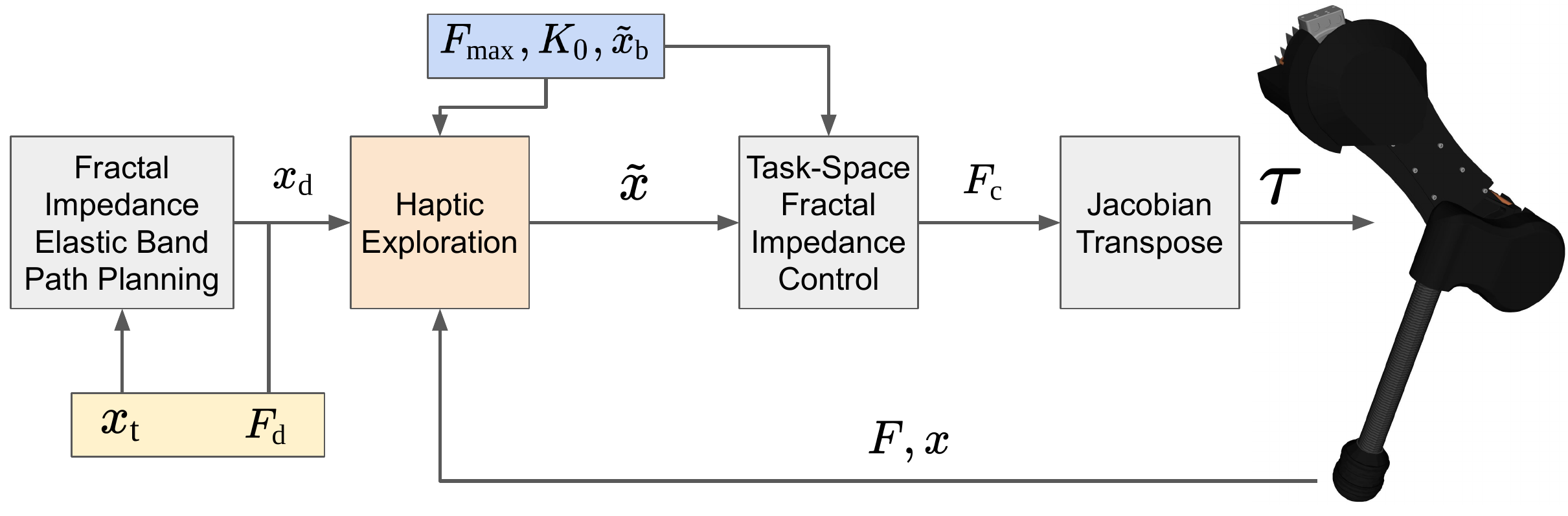}
    \caption{Overview of the system: A user provides desired task-space position and force $x_\text{t}$, $F_\text{d}$ (yellow) which are modulated through a passive elastic band controller to produce a target position $x_\text{d}$ at each time-step. User-tunable controller parameters $F_\text{max}$, $K_0$, and $\tilde{x}_\text{b}$ (blue) define the behaviour of the fractal impedance controller. The haptic exploration presented in this work (orange) outputs an adjusted target $\tilde{x}$ and maximal force $F_\text{b,max}$ to be used in the passive Task-Space Fractal Impedance Control. The task-space force $F_\text{c}$ is then mapped via the transpose of the Jacobian to joint torques $\tau$ which are executed on the anthromorphic arm---a single leg of the ANYmal quadruped robot. Sensed task-space positions $x$ and contact force $F$ are used by the haptics to close a force-feedback loop in world frame.}
    \label{fig:overview}
\end{figure*}

A haptic task can be defined as an action that relies on the sense of touch for its completion or for achieving maximum dexterity of interaction \cite{tiseo2020bio}. As a consequence, haptics is often encountered when providing force feedback to an operator~\cite{babarahmati2020,Erder2014,Erder2015}. However, the development of prosthetic, humanoid robotics and the deployment of robots in unstructured environments has shown the importance of robust control architectures capable of adjusting the trade-off between interaction force and position online \cite{Lotti2020,xin2020,tiseo2020}. The controllers deployed in these applications are usually based on the Optimal Port-Hamiltonian approach \cite{Averta2020,dong2019adaptive,xin2020,forceImpTrajLearning}. The two extreme behaviours for this approach are the admittance controller, where the trade-off favours force tracking, and the impedance controller where the trade-off favours position tracking \cite{impCtrl}. The need and the feasibility of a hybrid solution are documented in literature since the mid-nineties of the last century when the parallel force/position controller was proposed \cite{Siciliano1996}. However, the stabilisation of such a method required clear boundaries for the transition between different behaviours which is not always possible to obtain in unstructured environments \cite{Siciliano1996}. Nowadays, the impedance approach is used when robustness to potential perturbation which is undetected by the force sensor needs to be counteracted, usually in legged robotics---for instance early contact during locomotion in challenging terrain or for push recovery. Optimal admittance control prioritises interaction through the force sensor, which is typically used in industrial application and prosthetics. In rehabilitation robotics and exoskeletons, both approaches can be found, and the application mainly drives the choice.

Optimised Port-Hamiltonian controllers have been proposed to make the robots safe for interaction \cite{xin2020,Averta2020,TankBasedApproachImpCtrlVarStiff}. They drive robots using an equivalent mechanical system that guarantees the robustness of interaction by trading-off tracking accuracy and interaction force. Still, they need to rely on optimisation algorithms to guarantee accurate tracking and desired interaction force at the same time \cite{xin2020,Averta2020,Angelini2019}. Meanwhile, admittance controllers use the desired force as a driving signal to generate a desired force at the interaction \cite{Erder2014,Erder2015,HumanLikeAdaptOfForceAndImp,forceImpTrajLearning}. The robustness of both architectures is contingent on accurate modelling of the interaction dynamics that might be difficult to obtain in complex scenarios, rendering this type of architecture not well suited for unstructured environments. This is confirmed by looking at the literature where it is clear that most contributions in recent years are mainly driven by the development of more complex models enabled by the drastic increase of computational capabilities \cite{manchester2020,Chatzinikolaidis2020,ferrolho2020,wolfslag2020}. Dealing with making and breaking contact is challenging for both admittance and impedance control because they rely on accurate contact modelling for stability. Passive variable impedance controllers that adjust the trade-off between force and tracking accuracy performed by impedance controllers online might provide a solution, as their passive nature guarantees the controller stability \cite{babarahmati2019}.

Passive controllers have often been identified as a solution to the robustness conundrum due to their guarantees of stability if the control signal power is within the robot mechanical characteristics \cite{tiseo2020bio, tiseo2020, dietrich2015passivation,babarahmati2019}. These controllers are generally impedance controllers and can be classified into the intrinsically passive and the passivised controllers. The first type is passive by definition. They do not require any additional component to guarantee passivity. The simplest example of this controller is a critically damped passive impedance controller (i.e., desired velocity equal to zero). The passivised controllers use a virtual spring to evaluate the non-conservative exchange with the environment to guarantee passivity, and exploit the Port-Hamiltonian representation to perform a line integral of this energy \cite{TankBasedApproachImpCtrlVarStiff,dietrich2015passivation,babarahmati2019}. Thus, they allow velocity tracking as long as there is energy in the reservoir. Moreover, their stability depends on the accuracy of the energy tracking that is related to the discrete integration of the non-conservative energy, leading to state drift for low-bandwidth controllers \cite{babarahmati2019}. 

The \emph{Fractal Impedance Controller} (FIC) is a recently proposed framework that is intrinsically passive~\cite{babarahmati2019}. The fractal attractor guarantees the controller's asymptotic global stability by redistributing the energy accumulated in the controller spring during divergence to converge at the desired state following a harmonic trajectory. In other words, the fractal attractor is a generalised algorithmic representation of a critically damped passive system. If we use a linear stiffness to define the potential energy of the FIC, its behaviour is equivalent to a critically damped passive impedance controller. However, differently from a passive impedance controller, it allows a more general impedance profile by indirectly defining the controller impedance through the desired force profiles as a function of the position error (eq. \eqref{FICRA}), as described in the next section. Such an approach makes the controllers more intuitive and allows the definition of adaptive non-linear impedance that can be modulated online for different tasks. Finally, since the controller has a conservative energy and imposes an upper-bound on the control command power, the FIC concurrently guarantees global stability as well as robustness to low-bandwidth and delays \cite{babarahmati2019,babarahmati2020,tiseo2020Planner}.

Recently, we have implemented an adaptive force/position control that performs an online haptic exploration on a single degree of freedom without requiring any knowledge on the environment, which mimics muscle behaviour \cite{tiseo2020bio}. To do so, we exploited the guarantee of global stability to overcome one of the main limitations of a traditional force/position controller that requires the identification of transition zones between the force- and position-driven control strategies. The experimental results showed that the controller could safely switch between the two modes without requiring a state machine to switch between control strategies.

\textcolor{\SelCol}{This manuscript extends the work to an articulated mechanism based on the three degrees-of-freedom (DoF) leg (\autoref{fig:intro}) of the ANYmal robot~\cite{Hutter2016} (ANYbotics, AG). The aim is to test if the haptic controller can generate motor synergies across multiple joints. This would allow the Haptic Fractal Impedance Controller (HapFIC) in \autoref{fig:overview} to be deployed to interact with humans and unknown environments. Having robust haptic interaction is beneficial both for virtual and real experimental applications. The capability to generate stable and robust haptic interaction can be exploited for designing more realistic virtual environments in the future, which are beneficial for rehabilitation therapies \cite{hayre2020virtual,ozen2020haptic}. We have chosen the simulation over a robot validation because it represents a greater challenge to the proposed architecture that does not benefit from the accessibility to a deterministic knowledge of the environment, and, differently than on a robot, it has to deal with the numerical stability of the physics simulator.}

\section{Haptic Fractal Impedance Controller}
\begin{figure}[t]
    \centering
    \includegraphics[width=.7\columnwidth]{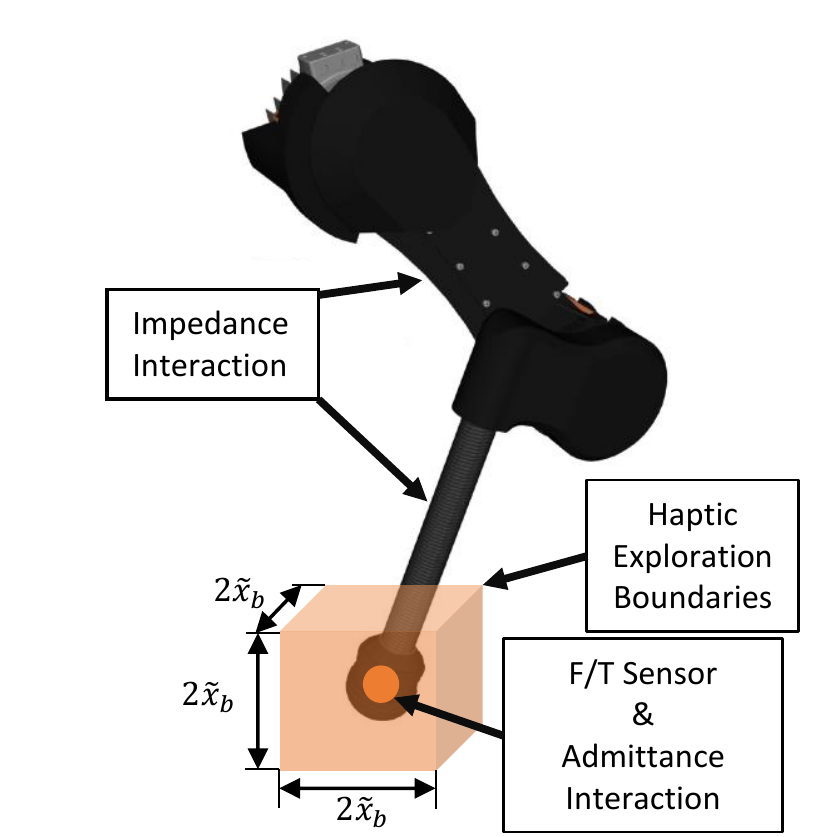}
    \caption{\textcolor{\SelCol}{Compound effect of the FIC and the Haptic module during interaction. The FIC handles all the physical interaction bypassing the force/torque sensor (F/T Sensor). Meanwhile, the interaction through the F/T Sensor is handled by the haptic module. The resultant admittance controller is bounded within the orange volume determined setting the $\tilde{x}_\text{b}$ of the FIC impedance profile, which can be interpreted as the desired position accuracy of the task.}}
    \label{fig:ControlInteraction}
\end{figure}

The proposed method (\autoref{fig:overview}) includes a planning architecture taking as input a desired target pose and generating an harmonic trajectory for the end-effector. The output of the planner is the input to the Haptic module that combines it with the desired force as well as the force feedback. \textcolor{\SelCol}{The Haptic module then modifies the planned trajectory to adjust the desired pose of the Task Space  Fractal Impedance Controller (TS-FIC) to generate the desired interaction behaviour. However, such admittance behaviour is bounded within the desired task precision that is controlled by setting the parameter $\tilde{x}_\text{b}$ in the FIC equations described in \autoref{sec:TS-FIC}. \autoref{fig:ControlInteraction} provides a graphical description of the interaction ports that the proposed method enables on the robot.}

\textcolor{\SelCol}{The FIC fractal attractor is central in guaranteeing stability, and its anisotropic force field is governed by the following equations during divergence (Div) and convergence (Conv), respectively \cite{tiseo2020bio,babarahmati2019}}.
\textcolor{\SelCol}{\begin{equation}
    \label{FICattractor}
    F_\xi(\tilde{x})=\left\{\begin{array}{cc}
        F_\text{c}(\tilde{x}),  & \text{Div} \\
        \cfrac{2F_\text{c}(\tilde{x}_\text{max})}{\tilde{x}_\text{max}}\left(\tilde{x}-\cfrac{\tilde{x}_\text{max}}{2}\right) & \text{Conv}
    \end{array}\right.
\end{equation}
where $\tilde{x}=x_\text{d}-x$ is the state error, $x_\text{d}$ is the desired state, $x$ is the current state, $F_\text{c}(\tilde{x})$ is a desired force profile, and $\tilde{x}_\text{max}$ is the maximum state error recorded at the beginning of the last convergence phase.}

\subsection{Harmonic Trajectory Planner}
The harmonic trajectory planner uses a Model Predictive Control (MPC) architecture to generate a harmonic trajectory between two points using a FIC, which was introduced in \cite{tiseo2020Planner} and extended in \cite{tiseo2020theoretical}. \textcolor{\SelCol}{The planner exploits the elastic field of the controller for generating smooth trajectories by integrating the acceleration generated by the FIC. To compute the accelerations, the stiffness $K$ is normalised by the system mass ($M_\text{d}$) and the maximum force is replaced by a maximum acceleration $A_\text{max}=2 v_\text{max}^2/d$ in every direction. Here, the maximum velocity ($v_\text{max}$) is computed by multiplying the desired velocity for the ratio between the average and peak velocities of the harmonic trajectories, which is $1.596$ as shown in \cite{tiseo2020theoretical}. Meanwhile, $d$ is the distance between the current desired position and the target. The equation of the desired trajectory is:}

\textcolor{\SelCol}{\begin{equation}
x_\text{d}=\iint_{t_0}^{t} \ddot{x}_\text{d}\left(t\right) dt^2
    \label{EqElasticBand}
\end{equation}
where $\ddot{x}\left(t\right)$ is determined via the FIC anisotropic behaviour for the divergence and the convergence to the target location.
\begin{equation}
       \begin{array}{l}
        \ddot{x}_\text{d}\left(t\right)=\left\{\begin{array}{ll}
                 \mathrm{sign}\left(\tilde{x}_\text{t}\right)\mathrm{min}\left(\cfrac{K}{M_\text{d}}\left|\tilde{x}_\text{t}\right|,a_\text{max}\right),& \text{Div}\\
                 \cfrac{2A_\text{max}}{\tilde{x}_{\text{T}0}}\left( x_\text{d}\left(t-1\right)-\cfrac{\tilde{x}_{\text{T}0}}{2}\right), & \text{Conv}
              \end{array}\right.
    \end{array}
    \label{EqElasticBand2}
\end{equation}}
\textcolor{\SelCol}{\noindent where $\tilde{x}_\text{t}=x_\text{t}-x_\text{d}\left(t-1\right)$ is position error from the desired location, $a_\text{max}$ is the acceleration a limit, $A_\text{max}$ is the acceleration computed at the maximum displacement reach in the previous divergence phase ($\tilde{x}_{\text{T}0}$).}

\subsection{Haptic Module}
The haptic module implements the search for the desired interaction based on the feedback received from the force sensor and refines the algorithm introduced in \cite{tiseo2020bio} by extending it to a multi-DoF system. The algorithm was modified by adding a bypass of the haptics when the desired force $F_\text{d}=0$. Further, a reset of the haptic search has been introduced to reinitialise the haptic search after losing contact. The reactive haptic search has been limited to a neighbourhood of the desired position for the end-effector, allowing a maximum pose error of $\tilde{x}_\text{b}$. If the search needs to be expanded beyond this bound, it can be done either through online tuning of this parameter or by adjusting the end-effector's desired pose. The full haptic exploration algorithm is shown in Algorithm~\ref{alg2}.

\begin{algorithm} 
\SetAlgoLined
  \SetKwData{Left}{left}
  \SetKwData{Up}{up}
  \SetKwFunction{FindCompress}{FindCompress}
  \SetKwInOut{Input}{Input~~}
  \SetKwInOut{Output}{Output}
 \Input{$F_{\text{d}}$, $K_0$, $\sigma$ , $F(t-1)$, $x_\text{d}$, $\Delta x(t-1)$, $\tilde{x}_\text{b}$, Reset}
 \Output{$x_\text{d}^\text{h}(t)$, $\Delta x(t)$}
  \BlankLine
  $\delta \tilde{x}_0=\frac{F_{\text{d}}}{K_0}$\\
  $\delta \tilde{x}_0^\text{h}=\sigma\delta \tilde{x}_0$\\
  \eIf{|$F_\text{d}(t)-F(t-1)|\ge \sigma  ~ \And ~ F_\text{d} \ne 0$}{
      
       \eIf{$|F(t-1)|\le \sigma$
           }{
                $\Delta x(t)=0$
           }{
                 $\Delta F=F_\text{d}(t)-F(t-1)$\\
                 $\Delta x(t)=\mathrm{clamp}(\Delta x(t-1) - \mathrm{sign}(\Delta F)\delta \tilde{x}_0^\text{h},-x_b, x_b)$
            }
    }{
        $\Delta x(t)=\Delta x(t-1)$\\
          \If{Reset~=~true
        }{
           $\Delta x(t)=0$
         }
    }
  $x_\text{d}^\text{h}(t)=x_\text{d}(t)+\Delta x(t)$ \\
\nonl   where:
\nonl   $t$ is the discrete time variable,\\
\nonl   $x_\text{d}$ is the displacement from the reference position that is expected when making contact with the environment,\\
\nonl $\tilde{x}_\text{b}$ is end-effector position error where the FIC force saturates to its maximum value\\
\nonl $\sigma=0.01$ scaling factor for the force scanning resolution.
    
\caption{Monodimensional Haptic Exploration} \label{alg2}
\end{algorithm}

\subsection{Task-Space Fractal Impedance Controller}
\label{sec:TS-FIC}
The Task-Space FIC is the lowest module in the control architecture that ensures stability of interaction with the environment. The chosen force profile has a single sigmoidal to the maximum force that encloses the linear impedance profile set around the desired pose, based on the formulation proposed in \cite{tiseo2020theoretical}. This force profile has the advantage that it can be easily adjusted and scaled compared with earlier formulations.
The force profile is fully described by
\begin{equation}
    \label{FICRA}
    \begin{array}{l}
         F_\text{c}=\left\{\begin{array}{ll}K_0\tilde{x}=
              K_0 \left(x_\text{d}^\text{h}(t)-x\right),& \tilde{x}\le 0.95\tilde{x}_\text{b} \\
              \cfrac{\Delta F}{2}\left(\tanh\left(\cfrac{\tilde{x}-\tilde{x}_\text{b}}{S\tilde{x}_b}+\pi\right)+1\right)+F_0,& \text{o/w}
              \end{array}\right.\\
    \end{array}
\end{equation}
where $K_0$ is the constant stiffness, $\tilde{x}$ is the end-effector pose error, $\Delta{F}=F_\text{max}-F_0$, $F_0=0.95K_0\tilde{x}_\text{b}$ and $S=0.1353$ controls the saturation speed. The value chosen for $S$ scales the hyperbolic tangent to saturate the force in the remaining $\SI{5}{\percent}$ of $\tilde{x}_\text{b}$.

The desired forces $F_\xi$ are then projected in joint space to generate the joint torque command:
\begin{equation}
\label{TorqueCommand}
    \tau=J(q)^\text{T}W
\end{equation}
where $J(q)$ is the geometric Jacobian and $W=[F_\xi~0_{3\times1}]$ is the desired wrench.

\textcolor{\SelCol}{In summary, the proposed architecture combines the impedance interaction of the FIC together with an admittance-like behaviour for the end-effector interactions, (\autoref{fig:ControlInteraction}).}

\section{Simulation Experiments}
A simulator for the robot has been developed using the multibody library in Simscape (Mathworks, Inc). The chosen solver algorithm is \texttt{ode45} with a step-size range of $[10^{-5},~10^{-4}]$ \si{\second}. For the proposed controller, simulation is more challenging than deployment on the real robot as the FIC stability has less stringent constraints compared to those posed by the physics simulator's numerical stability. However, on the physical system, the controller needs to be tuned to have a torque control command within the robot's band-pass. Methods for such tuning are established in \cite{babarahmati2019,tiseo2020,tiseo2020Planner}. On the other hand, model-based controllers rely on the accuracy of models and feedback information. 
The optimisation algorithms that state-of-the-art model-based controllers rely upon have a more stringent stability requirement than the ones required of physics simulators. As a result, model-based controllers can attain exceptional performance in simulation. At the same time, this implies that when deploying these controllers on the real robot, they exhibit limited robustness to model and feedback inaccuracies \cite{Angelini2019,xin2020,wolfslag2020}.
Crucially, incorrect assumptions about contact states and properties---for instance if a limb is in contact with the ground or whether it is slipping---can lead to numerical instability resulting in catastrophic failure. In order to deal with these challenges, commonly, approaches deploy complex state monitoring and recovery mechanisms \cite{Angelini2019,xin2020}. \textcolor{\SelCol}{Notably, our proposed approach does not require these mechanisms due to its passivity and ability to modulate the set point based on haptic feedback.}

The kinematic tree used in the simulation is an anthropomorphic 3-DoF arm \cite{siciliano2010robotics}, where the links' lengths are \SI{0.05}{\meter}, \SI{0.3}{\meter} and \SI{0.275}{\meter}, respectively. The masses of the links are all of \SI{1}{\kilo \gram}. A force sensor placed on the end-effector measures the interaction force with the environment. The friction and contacts are modelled using the Spatial Contact Force block in the multi-body library. Contacts are described in terms of stiffness, damping and a transition region parameter. It also includes a friction model, that uses constant static and dynamic friction coefficients while the critical velocity parameter mediates the transition. The contact parameters for rigid objects (i.e., robot, box and floor) are $K_\text{r}=10^{6}~\si{\newton\per\meter}$, $D_\text{r}=10^{4}~\si{\newton\second\per\meter}$ and \SI{1}{\milli\metre}. The contact parameters for the ball (i.e. soft object) are $K_\text{s}=10^{3}~\si{\newton\per\meter}$, $D_\text{s}=10^{2}~\si{\newton\second\per\meter}$ and \SI{1}{\milli\metre}. The friction parameters are $\mu_\text{s}=0.5$, $\mu_\text{d}=0.3$ and a critical velocity of \SI{1}{m\metre\per\second}, where applicable.

Three simulation experiments have been designed to evaluate the properties of the proposed architecture to i) verify the robot's performance in interaction robustness, ii) establish its capability in counteracting slipping, and iii) evaluate its interaction with deformable objects. These are all challenging scenarios in model-based control, where an accurate model of the contact is required to stabilise the system. Furthermore, we allowed a human user to adjust the target position and interaction forces online, via a user interface in all the simulations. \textcolor{\SelCol}{A fourth simulation experiment has been performed to evaluate the impact of a torque bandwidth of \SI{20}{\hertz} and a torque peak to \SI{40}{\newton} as per ANYmal specifications. A constant joint damping of \SI{11.46}{\newton\metre\second\per\radian} in the joint mechanical model is applied to simulate a non-ideal behaviour. These conditions have been analysed in the deformable object simulation, which contains both soft (i.e., a ball) and hard (ie., the floor) interactions.} 

\begin{figure*}[t]
\centering
    \centering
	\includegraphics[width=\textwidth]{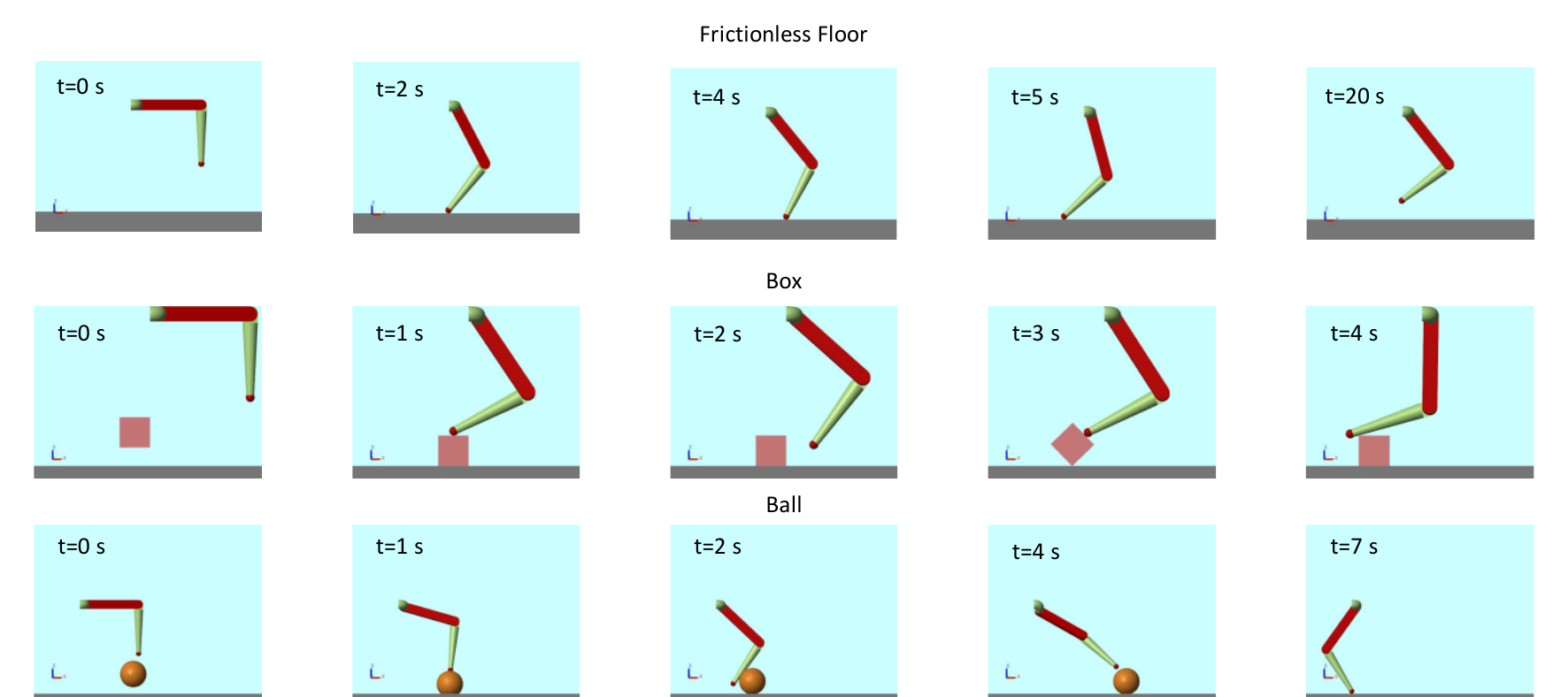}
	\caption{Salient snapshots from the simulation showcasing episodes in the three scenarios. The episodes shown for the friction-less floor are start, first contact with the ground, sliding forward, sliding backward and final configuration. The images shown for the box are start, first contact with the box, preparing to rotate the box, rotating the box, and making contact with the 'shank' while going to the final configuration. The episodes included for the interaction with the ball are start, pressing down, kicking with the 'shank', passing through the singularity and final configuration.}
    \label{fig:Snapshots}
\end{figure*}

The user also tuned the maximum force ($F_\text{max}$), the constant stiffness coefficient ($K_0$), and the search algorithm's reset condition to verify the stability during these parameters' online updates, which only allows serial updates of the different parameters. This is not an optimal update strategy, but it exposes the robustness of the system to extreme parameter values. For instance, this could be incompatible values of $K_0$, $F_\text{max}$ and $\tilde{x}_\text{b}$.
\textcolor{\SelCol}{For the three initial simulations, the initial algorithm parameter values are the same: $F_\text{max}=[150~150~300]$ \si{\newton}, $\tilde{x}_\text{b}=[0.025~0.025~0.025]$ \si{\meter} and $K_0=[6000~6000~12000]$ \si{\newton \per \meter}, and the reset condition is set to true. Meanwhile, the maximum force and the constant stiffness are adjusted to $F_\text{max}=[150~150~150]$ \si{\newton} and $K_0=[6000~6000~6000]$ \si{\newton \per \meter} during the fourth experiments to tune the Task-Space controller to the different hardware specifications.} 

The first experiment is an interaction with a friction-less floor. This implies that the controller cannot rely on the constraint generated by friction to stabilise its interaction with the floor. Therefore, the controller also has to generate constraints on the \textit{xy}-plane to generate the desired interaction along $z$. This problem is usually addressed in optimisation algorithms using the friction cones. This approach is extremely susceptible to the accuracy of the contact information, making it difficult to stabilise interaction when dealing with deformable bodies and non-linear dynamics \cite{xin2020,wolfslag2020}.

The second simulation experiment is the interaction with a box. It was divided into three sub-tasks: the first interaction is on the horizontal surface to generate controlled sliding and force interaction in the presence of friction. The second task is rotating the box to change the surface that is in contact with the ground. The third is evaluating robustness for unexpected interaction conditions at the end-effector as well as in other parts of the robot.

The last two simulation experiments involves the interaction with a deformable ball to validate the robustness of the proposed method in such conditions. Contacts with deformable bodies are usually difficult to model due to the contact conditions' volatility due to the distributed non-linear interaction between the two objects. As a result, it is difficult to generate the accurate model required by optimisation algorithms and controller to track the end-effector interaction. This experiment aims to verify the robustness of the proposed method in such a condition, which should be facilitated by the absence of an interaction model. Further, it is designed to highlight the robust and autonomous transition between contact and non-contact conditions.

\section{Results}
\begin{figure*}[!htb]
\centering
    \centering
	\includegraphics[width=\textwidth, trim=1.5cm 9cm 1.5cm 9cm, clip]{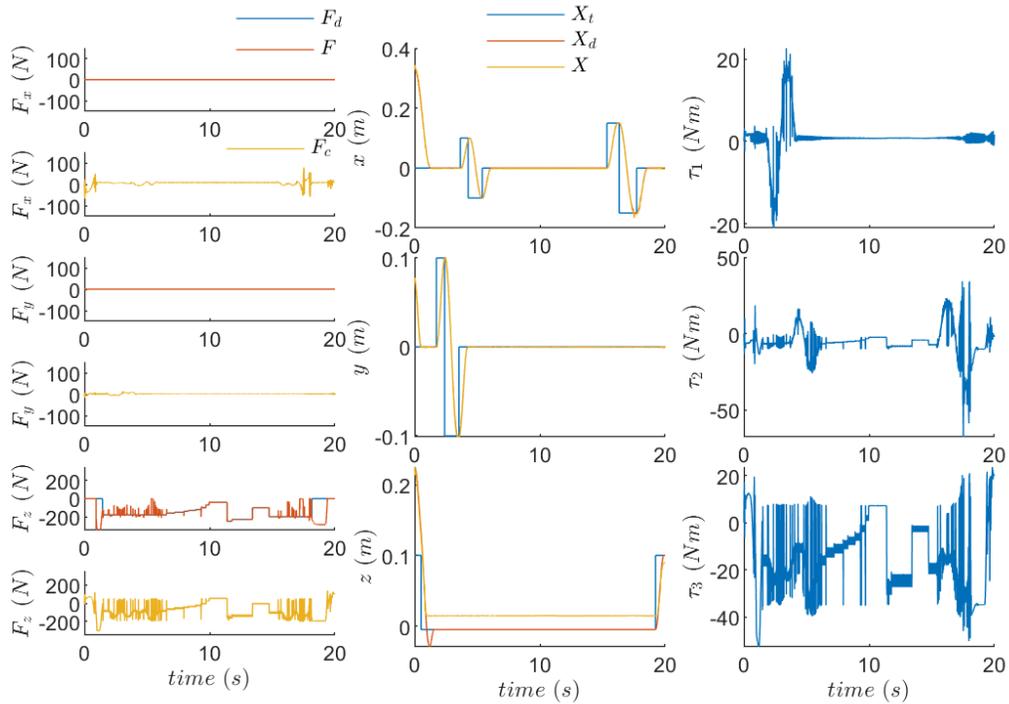}
	\caption{The interaction with the friction-less floor shows that the proposed method can handle slipping behaviour without relying on friction cones for stability. The FIC, as shown by the force command along \textit{x} and \textit{y} directions, automatically generates small compensatory signals to compensate the motion on the \textit{xy}-plane generated by the projections of the force normal to the surface. The position and force signals in the vertical direction show that the controller can track both force and position with the expected accuracy level, especially considering that all the sensors' feedback signals used in the controller are unfiltered.}
    \label{fig:FlatFloor}
\end{figure*}
\begin{figure*}[!htb]
\centering
    \centering
	\includegraphics[width=\textwidth,  trim=1.5cm 9cm 1.5cm 9cm , clip]{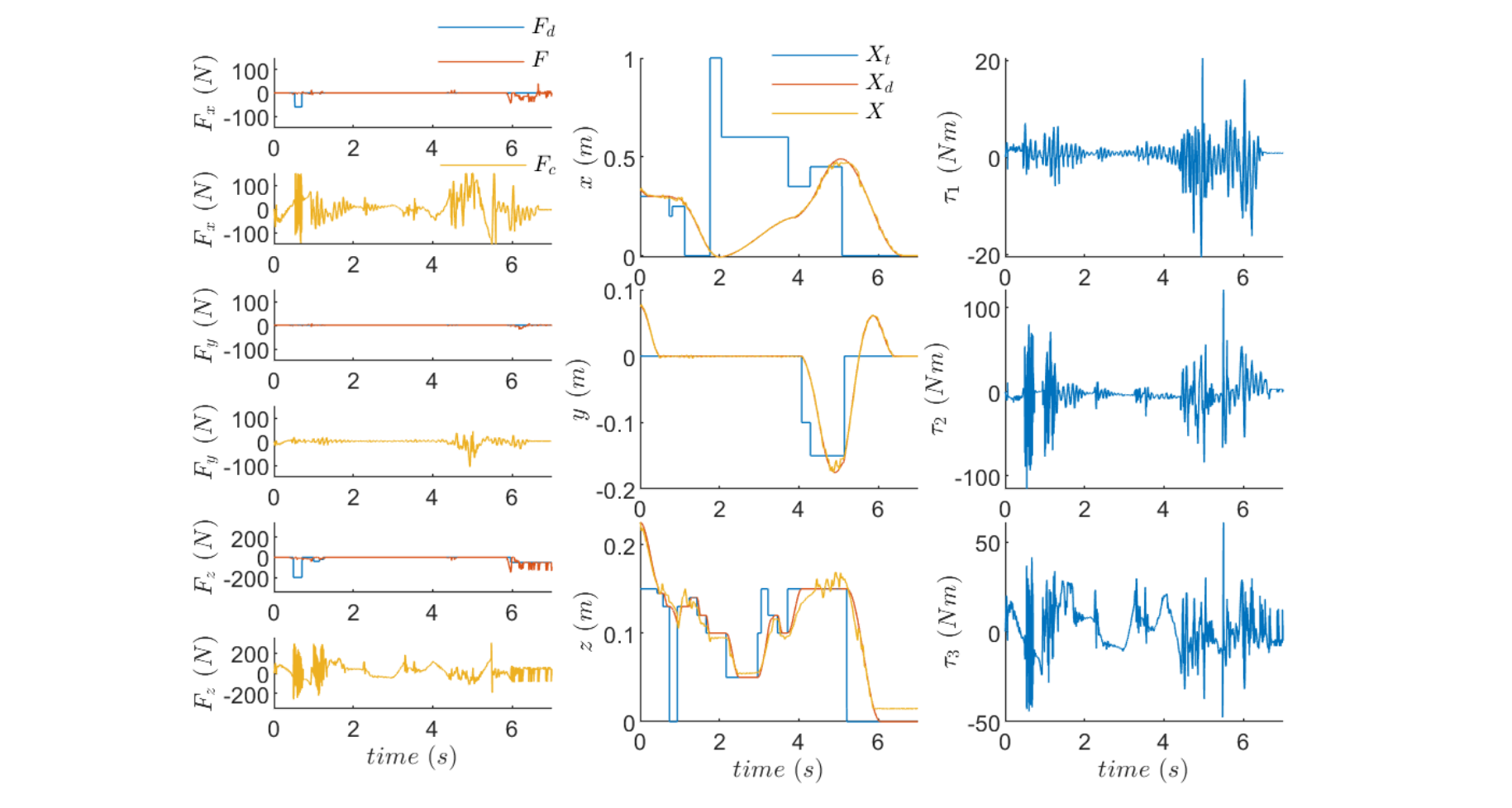}
	\caption{The interaction with the rigid box shows that the controller can at the same time perform force tracking at the end-effector while retaining the softness of interaction intrinsic to impedance controllers. This allows for tracking a force at the interaction point while having robustness of interaction to unexpected perturbations occurring in other locations. For example, this occurs when the shank makes unexpected contact with the box during the last action shown in the accompanying video.}
    \label{fig:BoxData}
\end{figure*}

\begin{figure*}[!htb]
\centering
    \centering
	\includegraphics[width=\textwidth,  trim=1.5cm 9cm 1.5cm 9cm, clip]{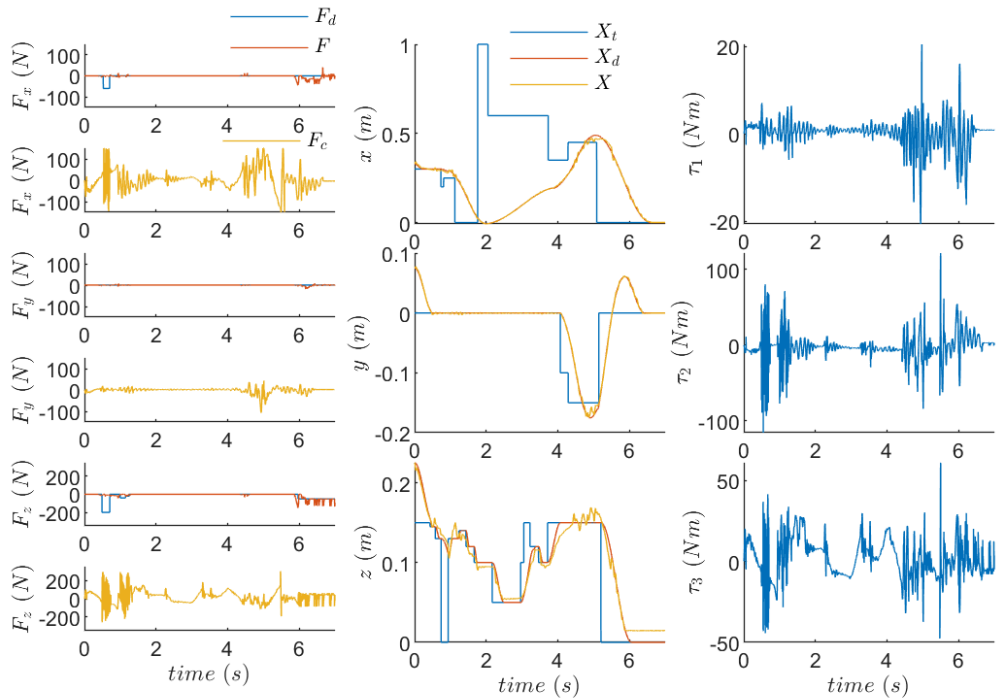}
	\caption{The interaction with a soft-ball indicates that it is possible to generate the desired interaction behaviour when dealing with deformable bodies, where determining friction and surface tangent vector is extremely challenging due to the contact condition's volubility. In particular, this demonstrates how the proposed method can autonomously make the trade-off between the contact and not-contact conditions without compromising stability.}
    \label{fig:BallData}
\end{figure*}

 
\autoref{fig:Snapshots} portraits five snapshots for each of the three simulations showcasing salient moments in the movements such as pushing, maintaining an object in equilibrium, and making and breaking contacts both at the end-effector and on the 'shank'. The simulation data shown in \autoref{fig:FlatFloor}, \autoref{fig:BoxData}, \autoref{fig:BallData} indicate that the forces cannot be accurately tracked in all the conditions. Especially when interacting with the friction-less floor, fluctuations in the contact force along the vertical direction are evident. This behaviour is probably connected to a trade-off with the impedance controller which takes over to generate the constraints required in order not to slip, because the force tracking error is reduced to a small vibration once the end-effector stops moving approximately at $t=\SI{6}{\second}$. Furthermore, this phenomenon is almost absent in the other two simulations.

The data also shows the trade-off between the haptic algorithm and the FIC: this time it is the controller taking over \autoref{fig:BoxData} at $t=\SI{2}{\second}$, that enables retaining a stable behaviour when both a non-zero desired force is set on the \textit{x} direction and the foot is not in contact. This would otherwise generate a divergence from the desired motion. The results of the simulation show that the robot is cable of tracking with an accuracy mostly constrained within the selected range of pose error,  $\tilde{x}_\text{b}$. However, the impedance controller takes over the authority when this happens and generates additional torques to compensate the external forces and minimises the tracking error.

\begin{figure*}[!htb]
\centering
    \centering
	\includegraphics[width=\textwidth,  trim=1.5cm 9cm 1.5cm 9cm, clip]{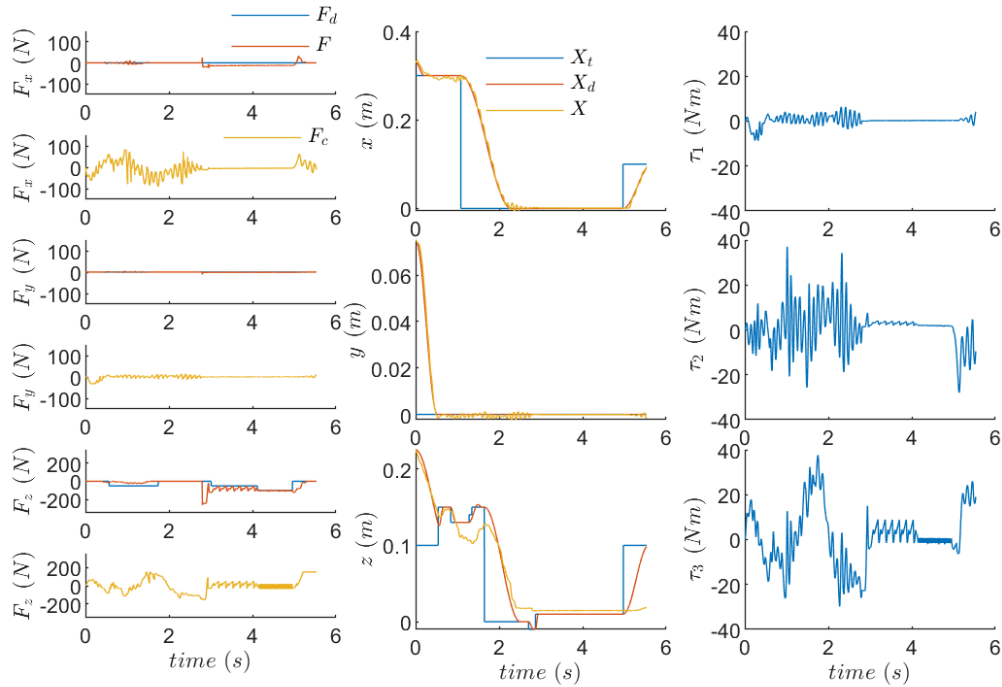}
	\caption{Introducing the actuation constraints reduces the hardware's responsiveness, but it does not jeopardise the robustness of the controller. The data indicate that the proposed control architecture can be adapted to meet the hardware band-pass, which seems to be the main factor in determining performance limitation in both trajectory and force tracking.}
    \label{fig:ActuatorsData}
\end{figure*}

\textcolor{\SelCol}{The fourth experiment data in \autoref{fig:ActuatorsData} indicate that the stability of interaction with hard and soft unknown dynamics is not affected by the introduction of the actuation limits and non-ideal joints mechanics (i.e., torque bandwidth and joint damping). However, the data also show a reduction in both trajectories and forces tracking performance that are consistent with the reduction of the system mechanical capabilities.}

\textcolor{\SelCol}{In all simulations, the tracking accuracy is consistent with the chosen $\tilde{x}_\text{b}$. However, the error increases beyond $\tilde{x}_\text{b}$ when the robot cannot generate sufficient forces to follow the desired motion, but these events do not jeopardise the system robustness that will recover as soon as possible as it can be observed in \autoref{fig:ActuatorsData} along the \textit{z}-axis. Meanwhile, $F_\text{c}$ is always within the constraint values selected at the controller initialisation ($F_\text{max}$). The end-effector position data also confirm the smoothness of the planned trajectories.}

\section{Discussion}
The HapFIC can generate robust interaction control while enabling the control of the interaction forces using the feedback from a force/torque sensor. It also enables the online adjustment of the trade-off between force and position tracking accuracy based on the tuning of the controller parameters $F_\text{max}$, $\tilde{x}_\text{b}$ and $K_0$. Furthermore, the proposed controller's properties guarantee that the magnitude of $F_\text{c}$ is upper-bounded by $F_\text{max}$ regardless of the haptic exploration algorithm. Meanwhile, the changing $\tilde{x}_\text{b}$ enables to extend or contract the haptic search domain, and $K_0$ also controls the search speed as can be seen in \autoref{alg2}. \textcolor{\SelCol}{It shall be noted that, as for any other FIC implementation, this architecture has to be calibrated to the mechanical characteristics of the system to ensure global stability. Within this work we followed the calibration procedure described in \cite{babarahmati2019}.}

\textcolor{\SelCol}{The proposed method enables haptic exploration within a predefined adjustable volume (\autoref{fig:ControlInteraction}) capable of stabilising and handling local high-frequency interaction. Doing so it decouples the geometrical complexity task from the system stability, as verified for path planning in non-convex domains in \cite{tiseo2020Planner}. This property allows handling tasks in highly non-convex scenarios via a geometric decomposition in quasi-convex sub-domains that can be handled by the proposed method. Furthermore, it implies that low-frequency admittance behaviour for longer movements can also be implemented by issuing adequate target sequences $x_{t}$ and producing an impedance causality admittance controller \cite{ott2009base}. A similar implementation has already been validated for controlling a 7-DoF collaborative robot (Franka Emika Panda) in the haptic teleoperation architecture presented in \cite{babarahmati2020} without any problem on the system stability even using communication delays up to \SI{1}{\second} and reduced communication bandwidth between the master and the replica robots.}

The HapFIC proved that it is possible to generate a controlled interaction with the environment without relying on numerical optimisation. This approach enables to reduce assumptions made about the contact, external dynamics and internal dynamics that would be otherwise required to control the robot. The additional robustness comes with the benefits of being more flexible in the interaction, significantly lower computation costs, and cross-form singularities. Nevertheless, there is the trade-off that the strategy is only locally optimal, while some optimisation algorithms guarantee global optimality. Therefore, optimisation-based methods are more suited for applications where there is a structured interaction with the environment, and the computational complexity is compatible with the task requirements.

Considering the properties discussed above, the HapFIC can be deployed when there is a need to adjust to sudden changes in the environmental conditions. Among these applications, there are legged locomotion, dexterous manipulation, human-robot collaboration and cohabitation. In these cases, traditional controlled architectures and optimisation algorithms have proven the feasibility in controlled conditions, but the complexity of the associated model often limits their deployment to more general scenarios. The computational simplicity of the HapFIC using the methodologies presented in \cite{tiseo2020,tiseo2020Planner,tiseo2020theoretical} opens the possibility for the development of distributed haptics along the entire body of the robot similarly to human skin. In legged locomotion and in other applications dealing with switching contacts and friction, this approach provides a more robust platform because it i) does not require \textit{a priori} knowledge of when the contact condition changes, ii) is robust to impact, and iii) is robust to sudden changes of friction. These are currently among the most daunting problems of such applications \cite{xin2020}.

Other fields where this controller might find application is in wearable robotics and understanding motor synergies. In wearable robotics, in order to maximise the systems' efficacy, it is essential to be able to interact with non-linear dynamics and to control interaction forces, while seamlessly being able to switch between different control parameters. For example, such architecture might control exosuits without relying on the non-linear dynamics of both the robot and the human biomechanics \cite{Lotti2020}. The biomechanic model is often coupled with bio-feedback to detect motor intention, increasing the costs and requiring an expert operator to be worn and used. \textcolor{\SelCol}{Regarding the motor synergies, the notion that Port-Hamiltonian controllers can be used to describe motor synergies is well-documented \cite{tommasino2017,Hogan2013}. However, the integration of distributed haptics has not yet been explored to the best of our knowledge. Therefore, our method also offers a more comprehensive model for motor synergies that might be explored in future work.}

\section{Conclusion}
The proposed HapFIC, and its ability to generate an adaptive parallel force/position control, have been validated in simulations across a range of challenging tasks. The results show that it can robustly interact with unknown dynamics and seamlessly switch between an impedance- and an admittance-like behaviour. The low computational costs of the controller make the proposed method relevant for many applications in fields such as robotics, rehabilitation and computational neuroscience.

\balance
\section*{Acknowledgements}
This work has been supported by the following grants: EPSRC UK RAI Hub ORCA (EP/R026173/1) and NCNR (EP/R02572X/1), and the EU Horizon 2020 project THING (ICT-2017-1).
\bibliography{main}
\bibliographystyle{IEEEtran}
\end{document}